\documentclass[a4paper,11pt]{article}
 \usepackage[margin=1in]{geometry}
\usepackage[square,numbers]{natbib}
\usepackage[utf8]{inputenc}
\usepackage{amsmath}
\usepackage{amsthm}
\usepackage{amsfonts}
\usepackage{bm}
\usepackage{thm-restate}
\usepackage{wrapfig}
\usepackage{subcaption}
\usepackage{float}
\usepackage{algorithm}
\usepackage[noend]{algorithmic}

\usepackage{color}   
\usepackage{hyperref}
\hypersetup{
    linktoc=all,     
    linkcolor=blue,  
}

\def\reals{{\mathbb R}}

\newcommand{\eps}{\epsilon}

\usepackage[group-separator={,}]{siunitx}
\usepackage{framed}
\usepackage{setspace}

\usepackage{todonotes}

\def\calL{\mathcal{L}}

\def\calN{\mathcal{N}}

\def\reals{{\mathbb R}}

\newcommand{\clip}{\mathrm{clip}}

\theoremstyle{plain}
\newtheorem{theo}{Theorem}[section]

\theoremstyle{definition}
\newtheorem{definition}[theo]{Definition}

\theoremstyle{remark}

\usepackage{dsfont}

\newcommand{\remove}[1]{}

\newcommand{\tr}{^{\mkern-1.5mu\scriptstyle\mathsf{T}}}

\allowdisplaybreaks

\begin{document}
\pagenumbering{arabic}

\title{
Large-Scale Differentially Private BERT
}
\date{}

 \author{
 Rohan Anil \\
 Google Research, Brain Team \\
 \small\texttt{rohananil@google\!.\!com}
 \and
Badih Ghazi\\
 Google Research \\
 \small\texttt{badihghazi@gmail\!.\!com}
 \and
Vineet Gupta\\
 Google Inc \\
 \small\texttt{vineet@google\!.\!com}
 \and
Ravi Kumar\\
 Google Research \\
 \small\texttt{ravi.k53@gmail\!.\!com}
 \and
Pasin Manurangsi\\
 Google Research \\
 \small\texttt{pasin@google\!.\!com}
 \and 
 }

\maketitle

\begin{abstract}
In this work, we study the large-scale pretraining of BERT-Large~\citep{devlin2018bert} with differentially private SGD (DP-SGD). We show that combined with a careful implementation,  scaling up the batch size to millions (i.e.,  mega-batches) improves the utility of the DP-SGD step for BERT; we also enhance its efficiency by using an increasing batch size schedule.  Our implementation builds on the recent work of \citep{subramani20}, who demonstrated that the overhead of a DP-SGD step is minimized with effective use of JAX \cite{jax2018github, frostig2018compiling} primitives in conjunction with the XLA compiler \cite{xladocs}.  Our implementation achieves a masked language model accuracy of 60.5\% at a batch size of 2M, for $\eps = 5.36$. To put this number in perspective, non-private BERT models achieve an accuracy of $\sim$70\%. 

\end{abstract}

\section{Introduction}
The widespread deployment of machine learning in recent years has raised serious concerns about the privacy of users whose data is used during the training process (see, e.g.,~\cite{kearns2019ethical}). These concerns are exacerbated by the well-known memorization behavior exhibited by deep neural networks~\cite{carlini2019secret} and in particular large language models~\cite{carlini2020extracting}. To mitigate these concerns, the framework and properties of differential privacy (DP)~\cite{dwork2006calibrating, dwork2006our} provide a compelling approach for rigorously controlling and preventing the leakage of sensitive user information present in the training dataset. Loosely speaking, DP guarantees that the output distribution of a (randomized) algorithm does not noticeably change if a single training example is added or removed; this change is parameterized by two numbers $(\epsilon, \delta)$---the smaller these values, the more private the algorithm. We refer the reader to Section~\ref{sec:prelim} for the formal definition of DP, and to~\cite{dwork2014algorithmic} for a thorough overview.

Motivated by these concerns, there has been a significant body of work on training private ML models. Notably, Abadi et al.~\cite{abadi2016deep} presented a generic recipe for training ML models with DP.  While their DP-SGD framework is quite robust as it applies to arbitrary neural network architectures, it faces two challenges that have significantly limited its practical deployment: (i) the gap between its accuracy and that of the best non-private methods can be significant, and (ii) the training time overhead (due to per-example gradient clipping) is considerable. Note that on simple tasks such as MNIST digit classification, the accuracy of DP models is not too far from that of non-private models. However, for more complex tasks such as CIFAR-$10$, the accuracy gap is very large, $\sim$25\% for reasonable privacy parameters. For even more complicated tasks such as CIFAR-$100$, the inefficiency of DP-SGD has for several years precluded the training of DP neural networks. These limitations have made the DP training of a complex language model such as the 
Bidirectional Encoder
Representation (BERT)~\cite{devlin2018bert} a daunting task.  

Very recently~\cite{hoory-etal-2021-learning} tackled the challenge of fine-tuning BERT with DP and relied on non-private pretraining on the combined Wikipedia and BooksCorpus \citep{ZhuEtAl2015bookcorpus} datasets.  In this work, we consider the task of pretraining a BERT-Large model with DP. We present an implementation of a variant of DP-SGD that, surprisingly, can train (relatively) quickly on state-of-the-art hardware and achieve good accuracy for this task.

\subsection{Contributions}
In this work, we establish a high accuracy baseline for DP BERT-Large pretraining. Our primary contributions are:
\begin{itemize}
  \item \emph{Negative interaction of naive DP-SGD with scale-invariant layers:} We discuss the importance of a large weight decay parameter (in Adam optimizer) and its interactions with layers that are scale-invariant individually and jointly. This insight allows us to tune hyper-parameters effectively, thereby achieving higher accuracies.   
  \item \emph{Mega-batches improve accuracy for DP BERT-Large:}  We demonstrate that scaling up the batch sizes to $\sim$2M improves the utility of every step of DP-SGD empirically. This batch size is \num{32}$\times$ larger than previously used for non-private training of BERT~\cite{nadogilmer21}. We achieve a masked language modeling accuracy of 60.5\%.
  \item \emph{Increasing batch size schedule improves training efficiency:} We show that an increasing batch size schedule can improve the efficiency of the training procedure while matching the accuracy of a fixed batch size schedule. We motivate our approach using a notion of gradient-SNR (signal-to-noise ratio). Our proof-of-concept experiments show up to $14\%$ reduction in the total number of examples visited to achieve the same accuracy.

\end{itemize}

\subsection{Related work}
Previous work \cite{rnndp2017} has trained recurrent language models with DP.
Previous work has also considered adaptive clipping in the context of DP-SGD \cite{andrew2019differentially, pichapati2019adaclip} as well as adaptive learning rates \cite{koskela2020learning}. In the non-private literature, increasing batch sizes have been considered, e.g., in \cite{smith2017don}. Finally, a significant speedup of DP-SGD was shown to be possible in \cite{subramani20} via large leaps in software \cite{jax2018github, frostig2018compiling, xladocs} for machine learning; this serves as our foundation for scaling pretraining for BERT.

\paragraph{Organization}
We start with some background in Section~\ref{sec:prelim}. The details of the algorithm are described in Section~\ref{sec:alg}. Our experimental setup and results are presented in Section~\ref{sec:experiments} and Section~\ref{sec:results}. We conclude with some future directions in Section~\ref{sec:conc}.

\section{Preliminaries}\label{sec:prelim}

Similar to most previous works on DP ML, we say that two datasets $X$ and $X'$ are \emph{neighboring}, if $X'$ results from adding or removing a single \emph{training example} from $X$. 
\begin{definition}[Differential Privacy \cite{dwork2006calibrating, dwork2006our}]\label{def:dp}
For any real numbers $\epsilon \geq 0$ and $\delta \in [0,1]$, a randomized algorithm $A$ is \emph{$(\epsilon, \delta)$-differentially private (DP)} if for every pair $X, X'$ of neighboring datasets and every subset $S$ of outputs of $A$, it is the case that $\Pr[A(X) \in S] \le e^\eps \cdot \Pr[A(X') \in S] + \delta$, where the probabilities are over the randomness in $A$.
\end{definition}

DP has seen significant interest in the literature due to its nice mathematical properties such as composition, post-processing, and group privacy (see, e.g., \cite{dwork2014algorithmic}).

\section{Algorithm}\label{sec:alg}

At a high-level, we use the DP-SGD algorithm \cite{abadi2016deep} with the Adam optimizer~\cite{kingma2014adam}.
Our choice of Adam follows the work of \citep{nadogilmer21}, who showed that tuning Adam works well up to large batch sizes of 65K.  As our goal is to establish a baseline for BERT pretraining with DP, we leave the investigation of higher-order methods \citep{anilpractical, amid2021locoprop} to future work. 

At each step of training, we randomly select a prespecified number of  examples, compute and clip their gradients and add appropriate noise to the average gradient to ensure privacy.  To compute the noise multiplier, we use the moment accountant (aka Renyi DP)~\cite{abadi2016deep,Mironov17} method implemented in TensorFlow Privacy\footnote{\url{https://github.com/tensorflow/privacy/tree/master/tensorflow_privacy/privacy/analysis}.}.  These noisy average gradients are then used to update the parameters via the Adam update \cite{kingma2014adam} rule with weight decay \cite{loshchilov2018fixing}.  See Algorithm~\ref{alg:DPSGD} for a self-contained complete description.

\begin{algorithm}[htb]
\caption{DP-SGD using Adam optimizer with weight decay\label{alg:DPSGD}}
\begin{algorithmic}
\REQUIRE Examples $x_1, \dots, x_n$, loss $\calL(\theta) = \frac{1}{n} \sum_{i=1}^n \calL(\theta; x_i)$. \\
\ENSURE number $T$ of steps, batch sizes $q_1, \dots, q_T$, clipping norm $C$, noise multiplier $\sigma$, momentum parameter $\beta_1$, second-moment averaging parameter $\beta_2$, weight decay $\lambda$, learning rate $\eta_t$.
\FOR{$t = 1, \dots, T$}
\STATE $B_t \leftarrow$ random $q_t$ training examples.
\STATE $g_t \leftarrow \frac{1}{|B_t|} \left(\calN(0, \sigma^2C^2 I) + \sum_{x_j \in B_t} \clip(\nabla\calL(\theta; x_j), C) \right)$
\STATE $m_t \leftarrow \beta_1 m_{t - 1} + (1 - \beta_1)g_t$
\STATE $v_t \leftarrow \beta_2 v_{t - 1} + (1 - \beta_2)g_t^2$
\STATE $\hat{m}_t \leftarrow m_t / (1 - \beta_1^t)$
\STATE $\hat{v}_t \leftarrow v_t / (1 - \beta_2^t)$
\STATE $\theta_t \leftarrow \theta_{t - 1} - \eta_t\left( \hat{m}_t / (\sqrt{\hat{v}_t} + \xi) + \lambda \cdot \theta_t \right)$
\hfill $\triangleright$ $\xi$ is set to $10^{-11}$
\ENDFOR
\RETURN $\theta_T$
\end{algorithmic}
\end{algorithm}

For the increasing batch size schedule, we modify the DP accounting procedure to be able to handle multiple batch sizes, in a straightforward manner. Specifically, for a fixed noise multiplier $\sigma > 0$, Renyi DP (RDP) order $\alpha > 0$, and batch size $q_t$, we modify the DP accounting procedure to compute $\eps_i(\alpha)$ such that the $t$th step is $(\alpha, \eps_t(\alpha))$-RDP. By composition for Renyi DP, the entire algorithm is  $\left(\alpha, \sum_{t=1}^T \eps_t(\alpha)\right)$-RDP, which can then be translated to an $(\eps, \delta)$-DP guarantee using the standard DP accounting procedure.

\section{Experimental Setup and Tuning} \label{sec:experiments}

We present results on pretraining a BERT architecture~\cite{devlin2018bert}, focusing on its larger variant, aka ``BERT-Large'', which is a transformer model \cite{vaswani17} containing $24$ transformer blocks with $1024$ hidden dimensions and $16$ self attention heads. It has $340$M parameters ($1.297$ GiB).  The training setup is a reimplementation of the official BERT codebase\footnote{\url{https://github.com/google-research/bert}} in the JAX framework \citep{jax2018github, frostig2018compiling} with the FLAX library \citep{flax2020github}. Our choice of JAX was motivated by the recent results of \citep{subramani20}, where they demonstrate that JAX features such as Just-In-Time (JIT)  compilation and compiler fusion result in low runtime overheads for the DP-SGD step and match or commonly exceeds the performance of various other frameworks. All experiments were carried out on Google TPUs \cite{JouYou17}, using the TPUv3-1024 configuration. 

\subsection{Pretraining dataset}
The pretraining dataset is the combined Wikipedia and Books corpus \citep{ZhuEtAl2015bookcorpus} datasets with 2.5B and 800M words, respectively. It consists of about 346M examples, each containing two sentences (389M unique sentences). To have a finite vocabulary and address out-of-vocabulary words gracefully, the words in the sentences are segmented into word-pieces \citep{sennrich2016neural}. There are 32K tokens in the vocabulary. Each pair of sentences has 128 tokens. 20 tokens from each example were replaced with masked tokens (15\% of the tokens). The objective of the model is to predict the masked tokens and which of the two sentences precedes the other. A typical pretraining model achieves 70\% in masked-language model (MLM) accuracy. After pretraining, the model parameters are used for fine-tuning on small amounts of data to solve specific language tasks \cite{glue}.

\subsection{Hyper-parameter tuning at 32K batch size}
\begin{figure}[htb]
\begin{center}
    \includegraphics[width=1\linewidth]{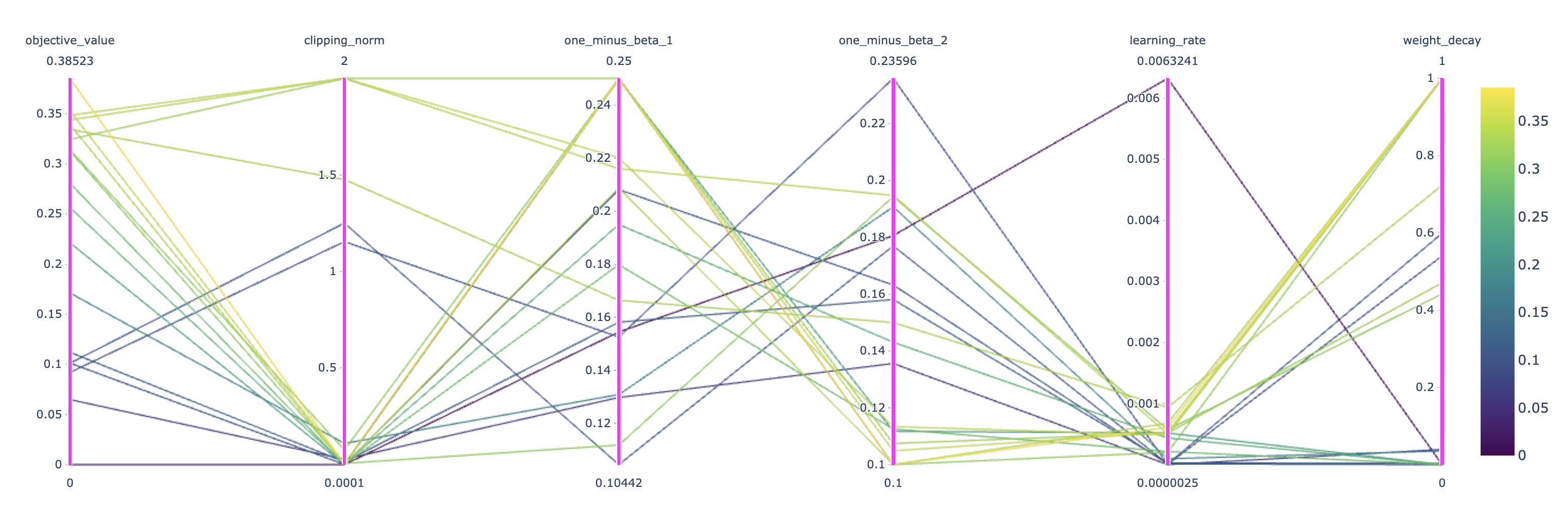}
    \caption{Tuning results at 32K batch size. Trained for 20K steps for 30 trials. Top trial yields an MLM accuracy of \textbf{38.5\%}.}
    \label{fig:vizier}
    \end{center}
\end{figure}

We tune hyper-parameters for Adam at a batch size of 32K and transfer the tuned hyper-parameters to all other batch sizes.  Note that our tuning step is \emph{not} done with DP; we leave this as an important future direction.
For tuning the hyper-parameters, we use a Bayesian optimization package and use a total of 30 trials. 
Hyper-parameter tuning spaces are listed in Table~\ref{table:search_space_bert}.  A linear learning rate warmup followed by a quadratic decay schedule is used for training. The tuning objective was to maximize the MLM accuracy over 10k examples following \citep{nadogilmer21}. Non-private training of BERT-Large reaches 70\% accuracy at 32K batch size in 14,063 steps. For all experiments, we train for 20k steps, with a 7.5K step warmup. We present results for tuning at 32K batch size for 20K steps in Figure~\ref{fig:vizier}. A key result is that a large weight decay $\lambda$ needs to be set for high accuracy.  We discuss this next.

\subsection{Scale-invariance and large weight decay}

The primary observation that led to larger search space for weight decay is that BERT-Large has several scale-invariant layers. A \emph{scale-invariant} layer is one in which increasing the norm of the weights in the layer by a positive scalar has no effect on the function output. However, the norm of the gradient for the weights of the layer shrinks as it is inversely proportional to the norm of the weights. This has been noted in the literature \cite{hoffer2019norm, Cho2017RiemannianAT, davody2020, heo2020adamp}, where \cite{heo2020adamp} propose a modification to Adam to handle these layers for non-private training, and \cite{davody2020} introduce batch normalization layers to make networks scale-invariant in order to improve accuracy, but do not consider the trainability issue of standard DP-SGD addressed here. 

To make this concrete, consider a fully connected layer with parameters $W \in \reals^{m\times n}$, where $Wx = s$. The gradient for the layer $G_t\in\reals^{m\times{}n}$ can be written via chain rule as $\nabla_s \ell(s_t,y_t) x\tr$. Under layer normalization, the preactivation vector $s$ is normalized to have zero mean and unit variance: $  \left(\frac{s - \mathbb{E}[s]}{\sqrt{\text{Var}[s]} + \xi}\right)$.
A key observation is that layer normalization makes the layer's output independent of the scale of $W$, i.e., multiplying $W$ by non-zero $\alpha \in \reals$ has no effect on the function output. With DP training, the Gaussian noise added to the gradient tends to increase the Frobenius norm $||W||_F$ of the weights over training which unintentionally shrinks the gradients, making training ineffective! Thus, to stabilize training, we set the weight decay parameter to be much larger than non-private training. This interaction is quite crucial for practitioners of DP to be aware of when applying DP to any neural network, as normalization layers such as layer-norm \cite{layernorm}, batch-norm \cite{batchnorm}, and weight-norm \cite{weightnorm} are common.

Notice that the difficulty in employing a more straightforward solution for the problem through gradient projection, i.e., projecting the component of the noise vector orthogonal to the weight vector, is that we need to infer the scale invariance property of the layers apriori, which is difficult. Moreover, for BERT-Large, the three types of embedding layers---wordpiece, positional, and token type---are scale-invariant together but not individually due to layer normalization being applied after aggregating these embeddings in the input layer. Thus, the heuristics from \cite{heo2020adamp} do not apply. Finally, handling the addition of standard DP noise to the scale-invariant layers efficiently by identifying them and including conjoint ones is an exciting avenue for further research.

\begin{table}[]
\centering
\begin{tabular}{c|c|l}
\hline

\textbf{Hyper-parameter} & \textbf{Range} & \textbf{Best trial} \\
\hline
$\eta$ (learning rate) & $ \lbrack 10^{-7}, 10^{-2} \rbrack $ & \num{6.0902e-4} \\ 
$1 - \beta_1$ (momentum parameter) & $ \lbrack 10^{-1}, 1.0 \rbrack $ &  \num{0.25} \\ 
$1 - \beta_2$ (second-moment parameter) & $ \lbrack 10^{-3}, 0.25 \rbrack $ &  \num{0.1} \\ 
$\lambda$ (weight decay) & $ \lbrack 10^{-1}, 1 \rbrack $ & \num{1.0} \\ 
$C$ (clipping norm) & $ \lbrack 10^{-6}, 2.0 \rbrack $ &  \num{3.2429e-3} \\ \hline
\end{tabular}
\caption{Hyper-parameter tuning search space, where the search is done in  logarithmic scale.  Learning rate schedule is a linear warmup followed by a quadratic decay. DP parameters $\epsilon$ is set to 5.36 and $\delta$ to \num{2.89e-9}.}
\label{table:search_space_bert}
\end{table}

\section{Experimental Results}
\label{sec:results}

In this section, we study the effect of the privacy parameter $\epsilon$ and fix a particular value of $\epsilon$ to aim for the highest attainable MLM accuracy, by changing the batch size.

\subsection{Varying privacy parameter $\epsilon$ }
\begin{figure}[h!]
\begin{center}
    \includegraphics[width=0.5\linewidth]{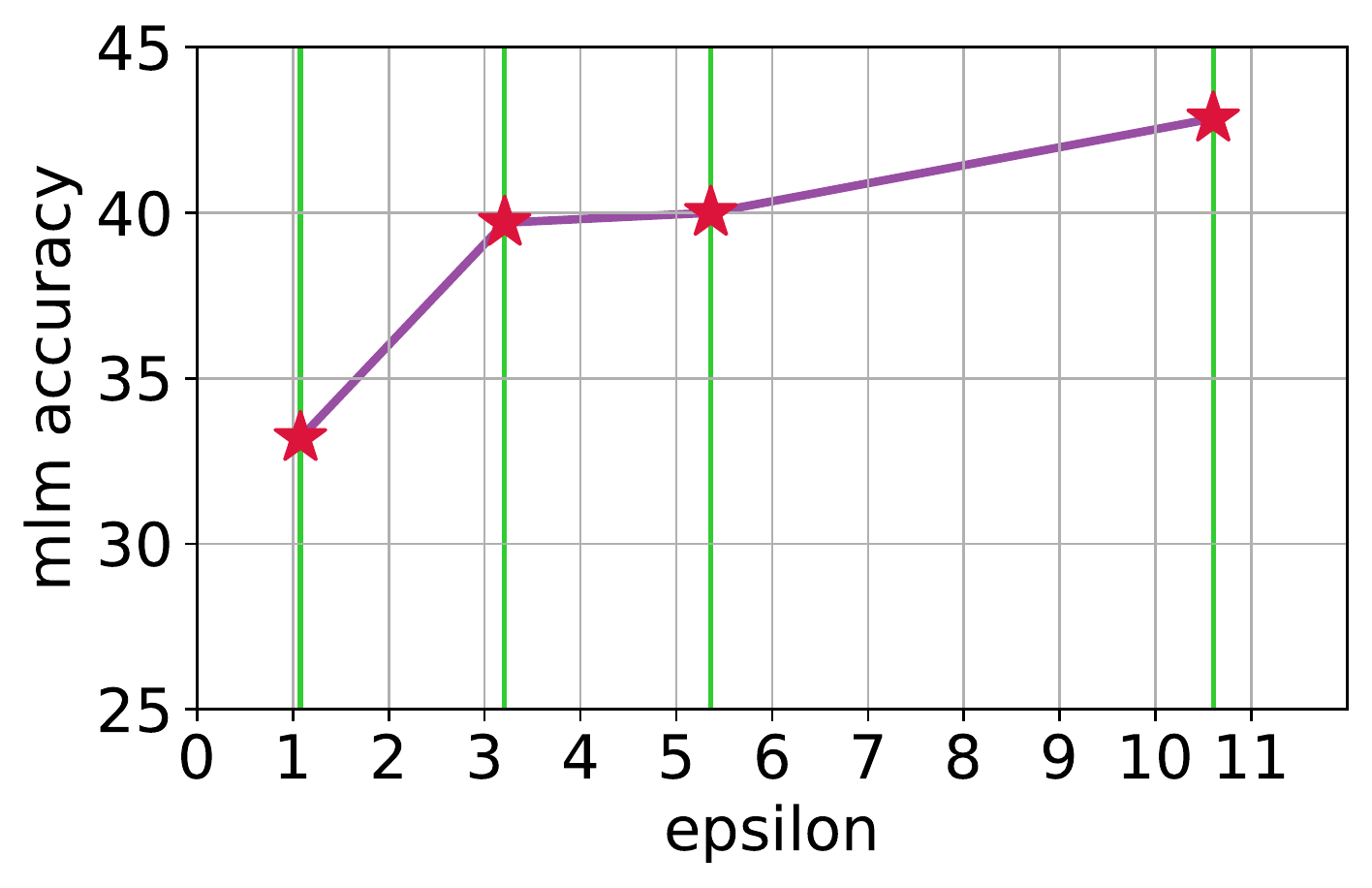}
    \caption{MLM accuracy tradeoff with varying $\epsilon$. We use $\epsilon$ of 5.36 for the rest of the experiments. Batch size is \num{65536} and trained for \num{20000} steps with \num{7500} steps of warmup.}
    \label{fig:epsilon_tradeoff}
    \end{center}
\end{figure}

In Figure~\ref{fig:epsilon_tradeoff}, we present the MLM accuracy results for varying the $\epsilon$ parameter. We set the batch size to \num{65536} and trained for 20K steps with a \num{7500} steps warmup. Hyper-parameters were transferred from the tuning experiments at \num{32768} batch size from Table~\ref{table:search_space_bert}. As expected, at $\epsilon=1.08$, the accuracy drops to 33.2\% while achieving up to 42.85\% at $\epsilon=10.6$. We identify a sweet spot of $\epsilon=5.36$ and use it for further experimentation; we also use
$\delta =$ \num{2.89e-9}, the reciprocal of the number of examples, throughout.

\subsection{Varying batch size under a fixed step budget}
We now study the effect of batch size on MLM accuracy.  In non-private training, typically batch size scaling is studied under a fixed epoch budget \cite{batchscience, nadogilmer21} rather than a step budget. In contrast, we carry out this study at a fixed step budget for two reasons: (i) our primary motivation is to establish a high accuracy private baseline, and (ii) fixed epoch budget study requires re-tuning hyper-parameters efficiently for very large batch sizes, which is still an open challenge for private training. 
Improving optimization efficiency of the training procedure is a natural next step, and along these lines, we propose an increasing batch size schedule that improves the efficiency and is described in the later sections.  (See Section~\ref{sec:conc} for  other possibilities for efficiency improvement.)  Notice that for all our experiments, we transfer the hyper-parameters from the 32K hyper-parameter tuning and naively increase the batch size while fixing $\epsilon$ to $5.36$.

\subsubsection{Effect of batch size on DP gradients} 
\begin{figure}[h!]
     \centering
     \begin{subfigure}[b]{0.40\linewidth}
         \centering
         \includegraphics[width=\linewidth]{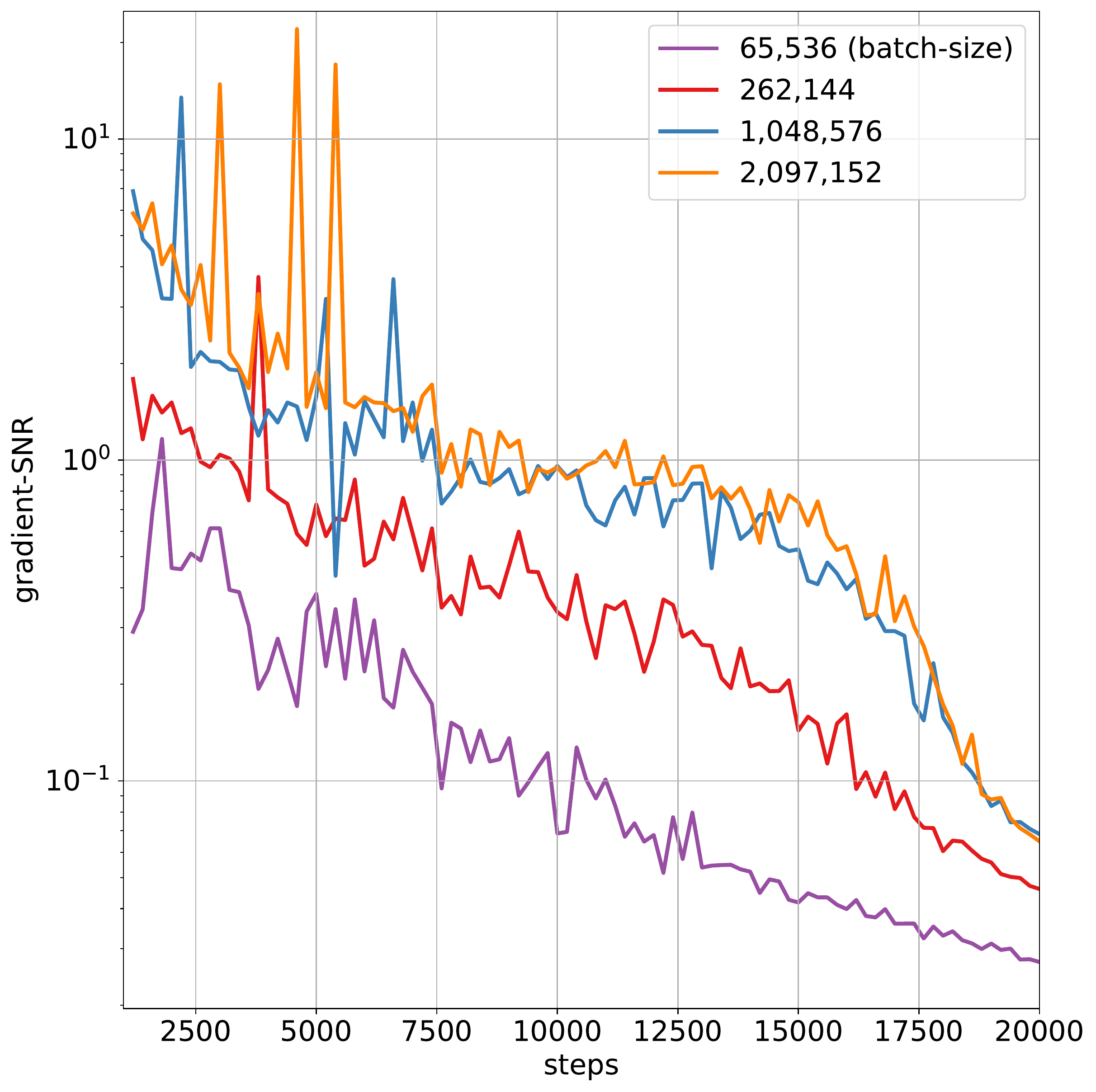}
         \caption{Gradient-SNR}
         \label{fig:gradient_snr}
     \end{subfigure}
     \begin{subfigure}[b]{0.40\linewidth}
         \centering
         \includegraphics[width=\linewidth]{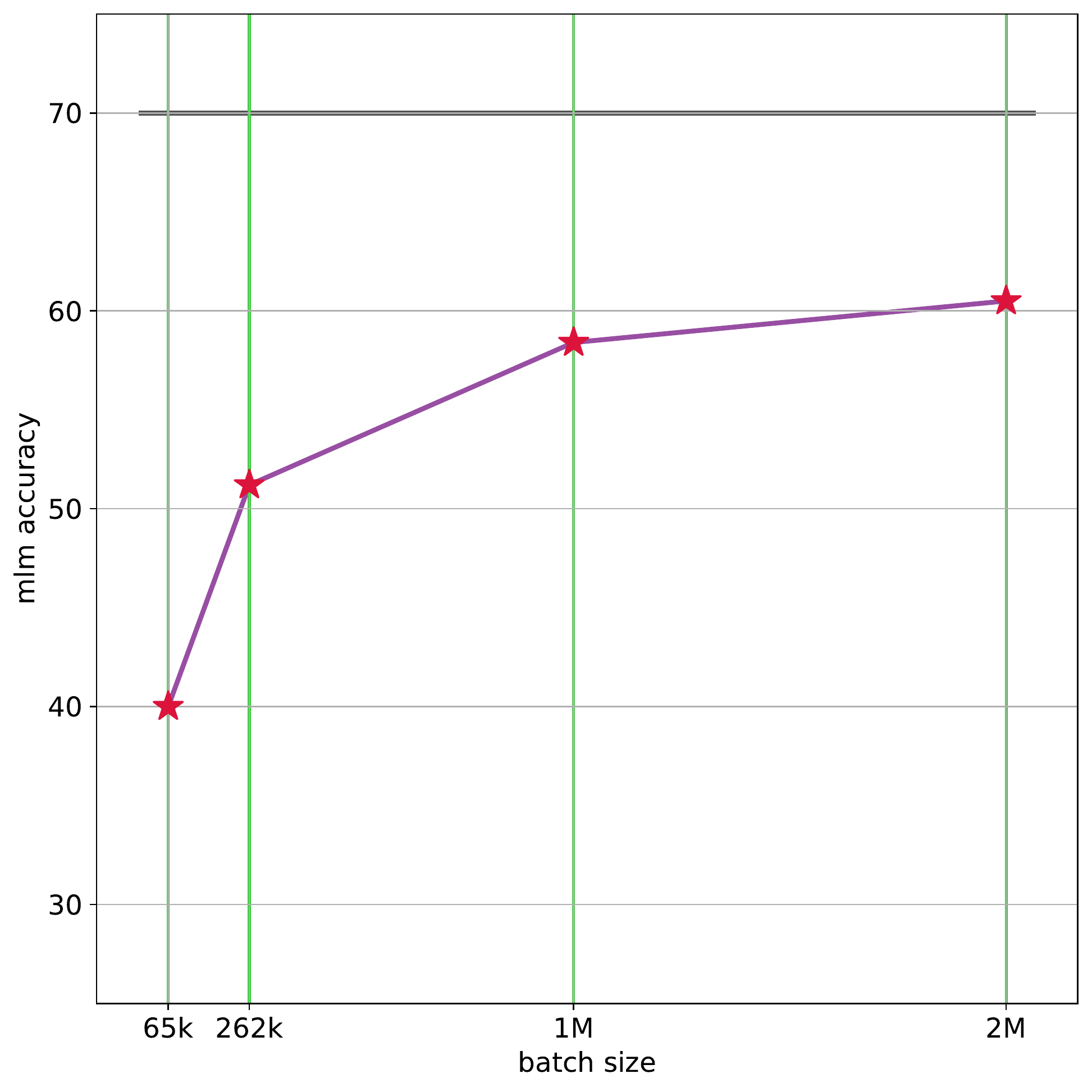}
         \caption{MLM accuracy}
         \label{fig:mlm_accuracy}
     \end{subfigure}
    \caption{Gradient-SNR and MLM accuracy at several batch sizes at a fixed step budget of 20K steps using hyper-parameters from Table~\ref{table:search_space_bert}. Larger batch sizes improve overall accuracy. Target solution quality is 70\% and is achieved by BERT-Large at modest batch sizes of 32K in \num{14063} steps.}
\end{figure}

As described in Section~\ref{sec:alg}, DP training relies on clipping individual gradients, then aggregating them, followed by the addition of a noise vector. We measure the ratio between the norms of the aggregated gradient of the network and the noise vector. We call this quantity gradient signal-to-noise ratio (gradient-SNR) and measure it over the training run; the results are presented in Figure~\ref{fig:gradient_snr} and Figure~\ref{fig:mlm_accuracy}. We observe that using a larger batch size yields favorable gradient-SNR through training and overall higher accuracy. Gradient-SNR can be seen to shrink as training progresses, which motivates the batch size schedule that we discuss next.

\subsubsection{Batch size schedule improves efficiency}
\begin{figure}[h!]
     \centering
     \begin{subfigure}[b]{0.40\linewidth}
         \centering
         \includegraphics[width=\linewidth]{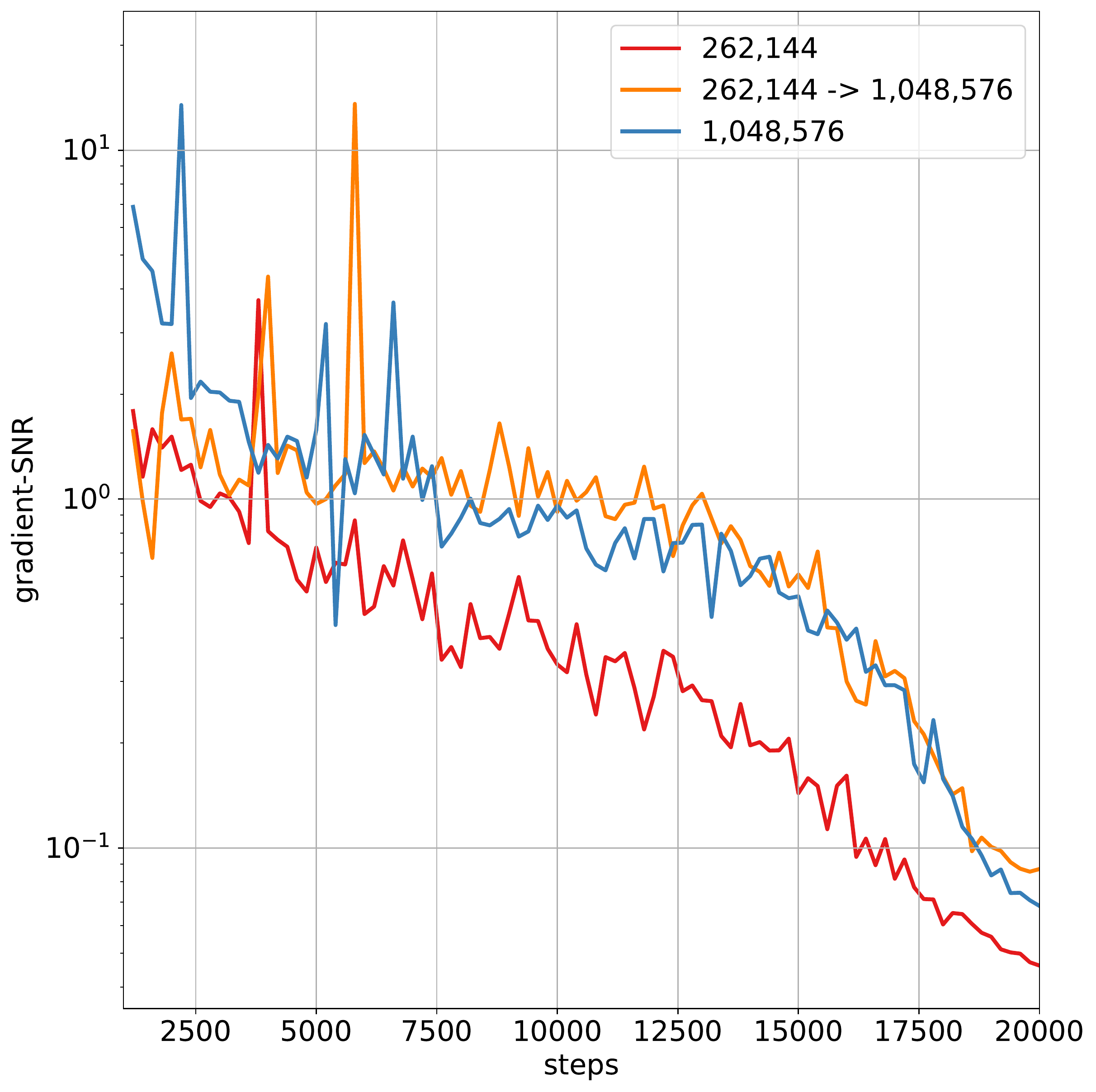}
         \caption{Gradient-SNR}
         \label{fig:dynamic_gradient_snr}
     \end{subfigure}
     \begin{subfigure}[b]{0.40\linewidth}
         \centering
         \includegraphics[width=\linewidth]{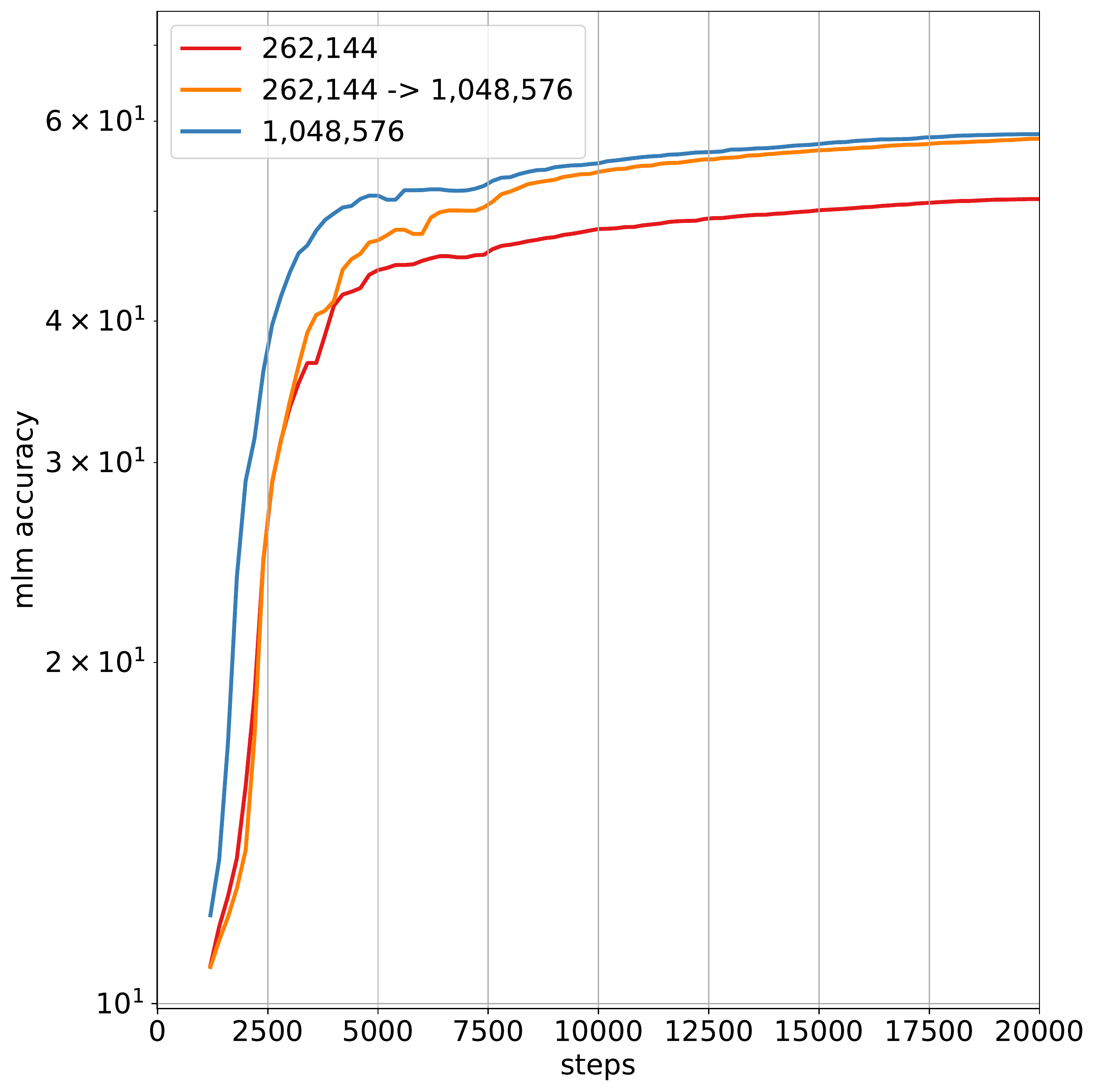}
         \caption{MLM accuracy}
         \label{fig:dynamic_mlm_accuracy}
     \end{subfigure}
    \caption{Gradient-SNR and MLM accuracy for fixed batch sizes 262K, and 1M, and increasing schedule starting at 262K to 1M over 7500 steps.}
\end{figure}

To improve the efficiency of training, we propose an increasing batch size schedule. As seen in Figure~\ref{fig:gradient_snr}, the gradient norm decreases over time, and the noise starts dominating, which leads to slower convergence of DP training. Based on this insight, we devise an increasing batch size schedule from $262,144$ (262K) to $1,048,576$ (1M) over 7.5K steps. Every one-quarter of the 7.5K steps, we increase the batch size by ~196k examples, reaching 1M at 7.5 steps. Overall we observe improved efficiency in training as seen in Figure~\ref{fig:dynamic_gradient_snr} and match the accuracy of fixed batch size training in Figure~\ref{fig:dynamic_mlm_accuracy}. This technique reduces examples seen in training by 14\%. We note that increasing batch sizes have been proposed in the literature, e.g., in \cite{smith2017don} where the technique is used as a substitute for decreasing the learning rate. In contrast, our proposal is motivated by the specific DP-SGD setting, i.e., the fact that the norm of the gradients tends to decrease as training progresses, so naturally, increasing the batch size allows us to improve gradient-SNR.

\subsection{On scaling up to mega batch sizes} 
The primary bottleneck with data-parallel training at large batch sizes ($>$65k) for BERT-Large is memory. Recall that \citep{you2019lamb} resort to using smaller batch sizes when training with longer sequence lengths (512) when using equivalent amount of hardware resources used in this work. As it is empirically clear that increasing batch sizes improves gradient-SNR and thereby improves the overall accuracy of the model, we handle mega-batch sizes by simply using gradient accumulation across examples to form large batches. JAX \cite{jax2018github, frostig2018compiling} makes it straightforward to implement this functionality by looping over batches of examples, and accumulating gradients obtained via a \texttt{jax.lax.fori\_loop} and \texttt{jax.vmap} (vectorized map). Moreover, larger batches are generally beneficial when training with slow interconnects, as they amortize the cost of gradient reduction. Increasing batch size is useful as long as the progress per example (utility) does not shrink. Another avenue is in the training of larger models ($>$5B parameters) that requires model parallelism. In this case, weights are split across devices, and efficiency is improved via pipeline parallelism where smaller batches are used to pipeline computations (forward, backward, update) across devices, as described in \cite{gspmd, gshard}. Integrating DP-SGD in that setup is the natural next step when training much larger models than BERT-Large. Another interesting question is how accuracy differs across various sizes of BERT models, which we leave for future work. To conclude, we use the gradient accumulation to scale up the batch size to \num{2097152} (2M), which is 32$\times$ larger than previously reported in the literature for non-private training, and obtain an MLM accuracy of \textbf{60.5\%} with DP.

\section{Conclusions}\label{sec:conc}

We build off the recent advances in software \cite{xladocs, jax2018github, frostig2018compiling} and hardware \cite{JouYou17} and establish a baseline for BERT-Large pretraining with DP.  We achieve high accuracy for the model by scaling up the batch size to millions of examples (mega-batches) and using additional insights such as improving the trainability of networks under normalization layers and measuring the gradient-SNR metric.  We proposed a proof of concept batch size increasing schedule and demonstrate an efficiency improvement. Another interesting direction is to improve the efficiency of DP training further by leveraging recent advances such as higher-order methods \cite{anilpractical, gupta2018}, local loss optimization \cite{amid2021locoprop}, scaling techniques that exploit increased parallelism: such as online distillation \cite{onlinedistillation} and ones that use lower memory \cite{shazeer2018, anil2019}, automatic tuning of hyper-parameters meta optimization \cite{egdd}, more efficient architectures such as MLP-Mixer, and F-NETs \cite{mixer, lee2021fnet} for longer sequences, multi-step training (data echoing) at mega-batch sizes \cite{choi2019faster, laggard}, loss functions that are robust to label noise \cite{bitemp}. We leave the study of the memorization properties of the DP trained model \cite{carlini2020extracting, memorization} as well as testing its generalization ability \cite{vitalylongtail} as future work.

\section{Acknowledgements}
We thank Ehsan Amid for feedback on the manuscript, Anselm Levskaya, James Bradbury, and Avital Oliver for helpful suggestions on the implementation. We also thank Aurko Roy, Ashish Vaswani, and Yonghui Wu for insightful discussions.

\bibliography{refs}
\end{document}